\documentclass[10pt, a4paper]{article}

\usepackage{lrec-coling2024} % this is the new style
\usepackage{times}
\usepackage{latexsym}
\usepackage{graphicx}
\usepackage[T1]{fontenc}
\usepackage[utf8]{inputenc}

\usepackage{microtype}

\usepackage{inconsolata}
\usepackage{xspace}

\usepackage{comment}
\newcommand*{\yoruba}{Yor\`ub\'a\xspace}

\title{Mitigating Translationese in Low-resource Languages: The Storyboard Approach}

\name{Garry Kuwanto$^{1}$, Eno-Abasi Urua$^{2}$, Priscilla Amuok$^{\dagger}$, \\ {\bf \large Shamsuddeen Hassan Muhammad$^{3\dagger}$, Anuoluwapo Aremu$^{\dagger}$, Verrah Otiende$^{\dagger}$}, \\ {\bf \large Loice Nanyanga$^{\dagger}$, Teresiah Nyoike$^{\dagger}$, Aniefon Akpan$^{2}$, Nsima Udouboh$^{2}$}, \\ {\bf \large Idongesit Archibong$^{2}$, Idara Moses$^{2}$, Ifeoluwatayo Ige$^{4\dagger}$, Benjamin Ajibade$^{\dagger}$}, \\ {\bf \large Olumide Awokoya$^{\dagger}$, Idris Abdulmumin$^{5\dagger}$, Saminu Mohammad Aliyu$^{6\dagger}$}, \\ {\bf \large Ruqayya Iro$^{\dagger}$, Ibrahim Said Ahmad$^{7\dagger}$, Deontae Smith$^{8}$, Praise-EL Michaels$^{8}$}, \\ {\bf \large David Ifeoluwa Adelani$^{9\dagger}$, Derry Tanti Wijaya$^{1,10}$, Anietie Andy$^{8}$}}

\address{$^{1}$Department of Computer Science, Boston University\\
$^{2}$University of Uyo, Nigeria,
 $^3$Imperial College London,
 $^{4}$Rochester Institute of Technology\\
 $^5$Data Science for Social Impact Research Group, University of Pretoria,
 $^6$Bayero University, Kano\\
 $^7$Institute For Experiential AI, Northeastern University \\
 $^{8}$Department of Electrical Engineering and Computer Science, Howard University\\
 $^{9}$University College London,
 $^{10}$Monash University Indonesia,\vspace{2mm}
 $^{\dagger}$Masakhane \\
\{gkuwanto,wijaya\}@bu.edu, \{deontae.smith,praise-el.michaels\}@bison.howard.edu\\
 anietie.andy@howard.edu}

\abstract{
Low-resource languages often face challenges in acquiring high-quality language data due to the reliance on translation-based methods, which can introduce the translationese effect. This phenomenon results in translated sentences that lack fluency and naturalness in the target language. In this paper, we propose a novel approach for data collection by leveraging storyboards to elicit more fluent and natural sentences. Our method involves presenting native speakers with visual stimuli in the form of storyboards and collecting their descriptions without direct exposure to the source text. We conducted a comprehensive evaluation comparing our storyboard-based approach with traditional text translation-based methods in terms of accuracy and fluency. Human annotators and quantitative metrics were used to assess translation quality. The results indicate a preference for text translation in terms of accuracy, while our method demonstrates worse accuracy but better fluency in the language focused.
 \\ \newline  }

\begin{document}

\maketitleabstract

\section{Introduction}

Low-resource languages pose significant challenges when it comes to acquiring high-quality language data for various applications, including language documentation, linguistic research, and machine translation \cite{kuwanto2023low}. Traditionally, data collection in these languages involves obtaining translations from higher-resource languages, such as English. However, this approach often leads to the introduction of a translationese effect, where the resulting sentences may be less fluent and natural including those produced by professional translators. 

Translationese refers to the linguistic phenomenon that occurs when translations exhibit characteristics that are not typical of the target language \cite{gellerstam-1986-translation}, as well
the use of more explicit and standardised constructions \cite{baker1993corpus} compared to original text. These characteristics can manifest as unnatural word choices, sentence structures, or even the adoption of foreign syntactic patterns. Translators, while highly skilled in bridging the language gap, often face challenges in recreating the nuanced meaning and linguistic nuances of the original text. As a result, the translated sentences may sound unnatural or stilted to native speakers, detracting from the authenticity and quality of the collected data.

Translationese has been widely recognized for its detrimental impact not only on machine translation tasks but also on other tasks involving cross-lingual transfer learning \cite{AmponsahKaakyire2022ExplainingTW,ni-etal-2022-original,artetxe-etal-2020-translation}. The presence of translationese introduces biases, diminishes fluency and naturalness, and ultimately affects the overall quality of the output. Previous research efforts have primarily focused on mitigating the translationese effect during the downstream phase, such as treating translationese as a different language \cite{riley-etal-2020-translationese}, applying embedding space projection \cite{yu-etal-2022-translate,chowdhury2022debiasing}, and utilizing paraphrasing techniques \cite{artetxe-etal-2020-translation,wein2023translationese}. However, these approaches have limitations, requiring additional annotation or modification of training data. Moreover, these methods primarily address translationese after its occurrence, rather than preventing or minimizing its presence during the data collection phase.
To the best of our knowledge, there has been no prior work specifically addressing the reduction of translationese artifacts during the data collection phase itself

In this paper, we address this challenge by introducing the use of storyboards \cite{burton2015targeted}, a common field linguist tool, in the data collection process. Our method leverages the power of visual stimuli to elicit more fluent and natural sentences from native speakers without the explicit influence of the source language text. Instead of providing sentences, we present native speakers with a storyboard consisting of images of scenes accompanied by their English sentences an hour before the annotation process, the process where annotators . During the annotation phase, participants are asked to describe the scene in the image, focusing solely on the visual content without access to the English sentences. The primary objective of our research is to examine whether storyboards can be an alternative method to data collection when improving fluency and reducing translationese bias are of interest. By removing direct exposure to source language text during annotation, we hypothesize that the resulting sentences will still hold the meaning of the original sentence while exhibiting improved fluency and naturalness. This approach has the potential to provide a more accurate representation of the target language, facilitating the development of higher quality language resources.

\vspace{-1mm}
\paragraph{Contributions:} 

We make three major contributions: (1) the collection of data in four typologically-diverse low-resource African languages (Hausa, Ibibio, Swahili, and \yoruba) in such a way that less translationese artifact arises, (2) the evaluation of the effectiveness of the storyboard approach in generating fluent and more natural sentences, and (3) to our knowledge, the first-ever parallel resource created data for Ibibio in non-religious domain.

In the following sections, we will delve into the details of our data collection method (Section \ref{sec:data}), describe the experimental design, evaluation process, and present the comprehensive results of our analysis (Section \ref{sec:experimental_design}). By combining qualitative evaluation by human annotators and quantitative metrics, we aim to provide an assessment of the effectiveness of our proposed method and its implications for data collection and the mitigation of the translationese effect.

\section{Related Work}
Data collection for low-resource languages has been a subject of ongoing research and development. Several initiatives have contributed to this field, such as the LORELEI project \cite{strassel-tracey-2016-lorelei} and the REFLEX-LCTL project \cite{simpson-2008-human}. These projects, conducted by the Linguistic Data Consortium (LDC), have released annotated corpora for multiple languages, addressing the need for linguistic resources in low-resource settings.

However, the predominant approach to data collection in low-resource languages still involves leveraging monolingual data from higher-resource languages and manually translating it. This is primarily because of the scarcity of monolingual and digital data available in the target language. However, this reliance on translation-based methods introduces challenges, such as the introduction of the translationese effect, where translated sentences may be less fluent and natural in the target language \cite{chowdhury2022debiasing}.

Translationese refers to the phenomenon in which translations exhibit linguistic characteristics that deviate from the typical patterns of the target language \cite{gellerstam-1986-translation}. 
% Despite their expertise in bridging language barriers, translators often face challenges in accurately capturing the nuanced meaning and linguistic nuances of the original text. 
The impact of translationese is particularly noticeable in the areas of syntax and grammar \cite{santos-1995-on}. Translations may exhibit unnatural sentence structures, lexical and word order choices that are influenced by the source language \cite{gellerstam1996translations}, adopt foreign syntactic patterns, as well as use more explicit and simpler constructions \cite{baker1993corpus}. This can result in sentences that sound unnatural or stilted to native speakers.

In the study by \cite{aranberri-2020-cantf}, an analysis of translationese was performed on the Spanish-Basque language pair. The researchers measured various linguistic features, including lexical variety, lexical density, length ratio, and perplexity. These measurements provided insights into the extent of translationese and its impact on the linguistic characteristics of the translated text. Similarly, \cite{university_of_tyumen_russia_translationese_2019} conducted a similar analysis focusing on English to Russian translation. 
In a different domain, \cite{bizzoni-etal-2020-human} conducted a study that compared translationese across human and machine translations from text and speech.

Multimodal translation tasks, such as the WMT18 multimodal task \cite{barrault-etal-2018-findings}, involve the use of both image and text during the translation process whether manually \cite{elliott-etal-2016-multi30k,ElliottFrankBarraultBougaresSpecia2017,barrault-etal-2018-findings} or automatically \cite{wijaya-etal-2017-learning,hewitt-etal-2018-learning,rasooli-etal-2021-wikily,khani-etal-2021-cultural}. During the manual annotation for these tasks, annotators are provided with both the source image and the corresponding source text, allowing for direct reference and alignment between the two modalities. However, this direct exposure to the source segment during translation may introduce the translationese effect, as observed in previous studies \cite{elliott-EtAl:2016:VL16}. For example, it has been observed that the lengths of translations in German are more similar to the lengths of the source English sentences than to the lengths of German image descriptions. This suggests that the presence of the source text can influence the resulting translations and potentially impact the fluency and naturalness of the target language output. In contrast, our storyboard-based data collection approach presents annotators with only the image, without the direct exposure to the source text.

% \subsection{Translationese}
% \subsection{Multimodal Data Collection}
% \subsection{Storyboard}

\section{Data and Data Collection}
\label{sec:data}
In this work, we propose a new method for collecting translations for low-resource languages, with the aim of reducing the influence of source (English) sentences during translation. To achieve this, we utilize a dataset comprising images depicting various scenes and their corresponding English descriptions. For each image and its associated English sentence, we engage two groups of native speakers for each target language. One group translates the English sentences, while the other group writes descriptions (in their respective languages) based on the visual content of the images. In this section, we provide a detailed description of the data and the data collection method employed in our study.

\subsection{English Sentence and Image Pair}
\begin{figure}[t]
\includegraphics[scale=0.4]{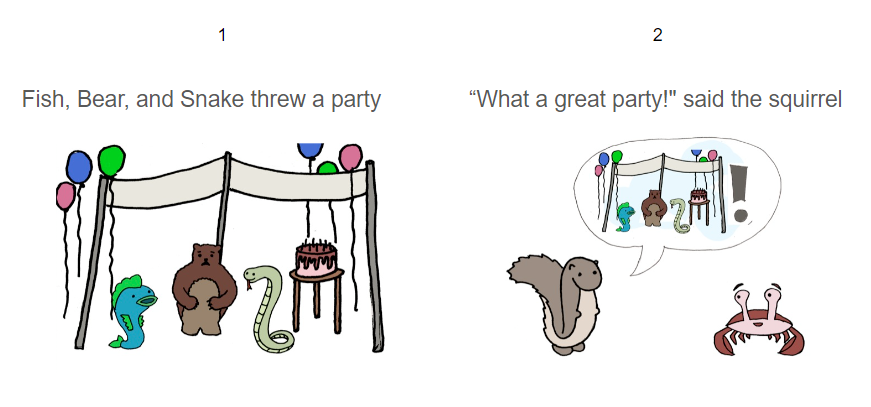}
\centering
		\caption{Example of English sentence and Image pair}
		\label{example_storyboard}
		\vspace{-1.5em}
\end{figure}

We obtain our dataset from the Totem Field Storyboards\footnote{\url{https://totemfieldstoryboards.org/}}, which provides a collection of storyboards consisting of sequential visual representations of stories in specific contexts, accompanied by corresponding English sentences describing the depicted scenes. An example of a sequence of English sentence and image pairs is shown in Figure \ref{example_storyboard}. For our research, we select 26 storyboards from the Totem Field Storyboards, each containing an average of 19 English sentence and image pairs.

\subsection{Translators and Focus Languages}
Our study focuses on four low-resource African languages: Swahili, \yoruba, Hausa, and Ibibio. For each target language, we engage a total of four native speakers. Among them, two native speakers are assigned to translate the English sentences, and the other two are shown the images and asked to write descriptions in their respective languages based on the visual content. % and one translator serves as a checker responsible for verifying the accuracy of the translations. In case of any errors found, the checker provides the correct translation instead of removing the erroneous one.

\subsection{Data Collection}
Considering our objective of determining and reducing the translationese effect in low-resource language data collection, we design the data collection process as follows: for each pair of English sentence and image in the storyboards, one group of translators focuses on translating the English sentence, while another group concentrates on translating the image. This separation allows us to examine the impact of English sentences on the resulting translations and evaluate the fluency and naturalness of the sentences generated solely based on visual content. By collecting data through this dual approach, we aim to obtain a more authentic representation of the target languages and mitigate the translationese effect. For each language, the translators were paid a total of 600 US dollars; which they shared equally among themselves.

\subsubsection{Control Group: Text Translation}

In our study, we include a control group that utilizes the traditional approach of text translation. This control group serves as a baseline for comparison and allows us to evaluate the effectiveness of our storyboard-based method in reducing the translationese effect.

For the control group, native speakers are provided with the English sentences from the storyboards and are instructed to translate them directly into the target languages. These native speakers possess proficiency in both the source and target languages. % and are experienced in translation tasks.

The translations generated by the control group represent the typical output obtained through traditional translation-based approaches. These translations are expected to exhibit characteristics of translationese, such as potential deviations from natural language usage, as translators may prioritize fidelity to the source text over fluency in the target language.

\subsubsection{Treatment Group: Storyboard-Based Translation}

In our storyboard-based translation method, we introduce a preparatory phase before the actual annotation process. Before annotating each storyboard, the annotators are grouped together in a meeting where they are given the opportunity to familiarize themselves with the storyboard and the corresponding English sentences. During this meeting, they can read through the storyboard and comprehend the context and content of each image.

After this reading session, we introduce a time gap of $\sim$1 hour before the annotation process begins. This time gap serves two purposes: (1) it allows the annotators to internalize the visual information from the storyboard, and (2) it minimizes the direct influence of the English sentences on their subsequent annotations.

During the annotation phase, the annotators are provided with the storyboards containing only the images, without any accompanying English sentences. They are instructed to focus solely on the visual content and describe in their respective target languages the scene that is being depicted in each image. This approach ensures that the annotations are driven primarily by the visual stimuli, encouraging the annotators to provide translations that capture the essence of the scenes portrayed in the images.

By removing the explicit exposure to the English sentences during the annotation process, our storyboard-based method aims to reduce the potential influence of the source language and the translationese effect. The annotators are encouraged to rely on their linguistic knowledge and cultural understanding to generate fluent and natural translations that are more aligned with the target language's usage patterns and stylistic conventions.

\subsection{Annotation Dataset}
The resulting annotation dataset consists of the source English sentences and their corresponding translations in four low-resource African languages: Swahili, \yoruba, Hausa, and Ibibio. The dataset includes both text translations and translations obtained through our storyboard-based method.

In total, we collected translations for 486 unique English sentences. The distribution of translations is shown in Table \ref{tab:dataset}, indicating the number of translations for each language and data collection method (text or storyboard). The number between text and storyboard translations differs slightly because we gave translators the option to provide alternative translations for each English sentences.

\begin{table}[ht]
\centering
\begin{tabular}{lcc}
\hline
\textbf{Language} & \textbf{Text Translation} & \textbf{Storyboard} \\
\hline
Hausa & 1154 & 968 \\
Ibibio & 887 & 883 \\
Swahili & 1334 & 1211 \\
\yoruba & 1448 & 1033 \\
\hline
\end{tabular}
\caption{Number of translations in the dataset}
\label{tab:dataset}
\end{table}

The dataset also includes additional information such as the title of the storyboard and the scene number associated with each translation. We will make the final dataset, along with the source English sentences, publicly available. An anonymized version of the dataset is also attached in the supplementary material for reference.

\section{Experimental Design}
\label{sec:experimental_design}

All computational experiments were conducted on a machine equipped with an Intel Xeon Gold 6226R CPU running at a clock speed of 2.90GHz and an NVIDIA RTX A6000 GPU.
\subsection{Human Evaluation}
\textbf{Fluency}, for our human evaluation setup, refers to the smoothness and naturalness of the translated sentence in the target language. A fluent sentence should not sound "foreign" or awkward, and should read as if a native speaker had originally written it in that language. Fluency captures the syntactic and grammatical correctness as well as the idiomatic usage of the language.

\textbf{Accuracy}, for our case, refers to the extent to which the translated sentence captures the meaning of the source sentence. An accurate or adequate translation should convey all essential information from the source text without adding, omitting, or distorting any content.

To assess the accuracy and fluency of the translated sentences obtained through the traditional text translation and our proposed storyboard-based data collection processes, we conducted human evaluation with native speakers who are also proficient in English as annotators. Annotators were assigned two tasks: one for accuracy evaluation and another for fluency evaluation. Each task involved comparing a pair of sentences, one from the text translation and the other from the storyboard-based collection process. It is important to note that the sentence pairs provided to the annotators for different languages were taken from the same storyboard scene  to ensure consistency and fair comparison across languages. For each task (accuracy and fluency), we randomly select 100 samples from the storyboard scenes and obtain sentence pairs.

To minimize bias and ensure reliable results, three human annotators were assigned to each task. Each annotator independently evaluated the sentence pairs and provided their preference based on accuracy or fluency. The preferences of the annotators were then tallied up to determine the overall preference for each translation approach.

After the evaluation, because we have 3 annotators for each language, the inter-annotator agreement was calculated using the Fleiss' Kappa statistic to measure the consistency of the annotators' preferences. A Fleiss' Kappa value of 1 indicates perfect agreement between the annotators, while a value of 0 suggests agreement equivalent to random chance. For our evaluation, the Fleiss' Kappa value was found to be 0.27 and 0.16 for accuracy and fluency respectively, indicating fair and slight level of agreement among the annotators respectively \cite{landis1977measurement}.

\subsubsection{Accuracy Evaluation}
For the accuracy evaluation task, annotators were presented with three sentences: the source English sentence, the sentence translated through text translation, and the sentence translated through the storyboard-based method. Their task was to choose which sentence they deemed more accurate for translating the English sentence; they can pick one sentence over the other, or both. The guideline was to "select which sentence is more adequate (i.e., more accurate) for translating the English sentence." An important criterion emphasized in the guideline was that a better translation should include as much content from the English sentence as possible, without adding information not present in the original sentence. The annotators were also asked to disregard the translations of named entities in their accuracy judgment. This was particularly relevant as, during the storyboard data collection process, the translators might not remember the exact names mentioned in the English sentence, while the translators for the text translation could reference the English sentence for accuracy.

To reduce bias, annotators were only provided with the sentences without any indication of which sentence was from the text translation and which was from the storyboard-based method. An example of the accuracy evaluation task for Hausa can be seen in Table \ref{tab:acc_example} where annotators would compare Sentence 1 and Sentence 2, selecting the one they considered more accurate, or both if they deemed the two sentences to be equally accurate.

\begin{table*}
    \centering
    \begin{tabular}{ll}
        \hline
        \textbf{English Sentence} & I bought shoes, a hat, a shirt, and pants \\
        \hline
        \textbf{Sentence 1} & Na siyo takalma, malafa, riga da kuma wando \\
        English Translation& \textit{I bought shoes, mattress, shirt and pants}\\\hline
        \textbf{Sentence 2} & Na sayo takalma, hula, riga da kuma yan kamfai \\
        English Translation& \textit{I bought shoes, a hat, a shirt and a few shoes}\\
        \hline
    \end{tabular}
    \caption{An Example of Accuracy Human Evaluation for Hausa. Annotators were asked to choose the sentence they deemed more accurate for translating the English sentence. They can answer with Sentence 1, Sentence 2, or both if they deemed both sentences to be equally accurate.}
    \label{tab:acc_example}
\end{table*}

\subsubsection{Fluency Evaluation}
For the fluency evaluation task, annotators were not provided with the source English sentence. Instead, they were presented with two sentences from the same storyboard scene: one sentence obtained through text translation and another obtained through the storyboard-based method. Their task was to select the sentence that they deemed to be more fluent and natural. The guideline was to "select which sentence is more fluent (i.e., more natural). A better sentence should be the one that is more natural and grammatical". 
The annotators were asked to focus on factors such as natural language usage, grammaticality, and overall fluency in making their decision. 

To ensure unbiased evaluation, the annotators were only provided with the sentences without any indication of which sentence was from the text translation and which was from the storyboard-based method. An example of the fluency evaluation task for Hausa can be seen in Table \ref{tab:flu_example} where annotators would compare Sentence 1 and Sentence 2, selecting the one they considered more fluent, or both if they deemed the two sentences to be equally fluent. 

\begin{table*}
    \centering
    \begin{tabular}{ll}
        \hline
        \textbf{Sentence 1} & Mary ta fita zuwa shago \\ 
        English Translation& \textit{Mary goes out to the shop}\\\hline
        \textbf{Sentence 2} & Ta fita siyayya watarana \\
        English Translation& \textit{She went out of the shopping}\\
        \hline
    \end{tabular}
    \caption{An Example of Fluency Human Evaluation for Hausa. Annotators were asked to choose the sentence they deemed more fluent. They can answer with Sentence 1, Sentence 2, or both if they deemed both sentences to be equally fluent.}
    \label{tab:flu_example}
\end{table*}

\subsection{Metric-Based Evaluation of Accuracy and Fluency}

In addition to human evaluation, we employ several metrics to support and complement the results obtained from the human annotators. These metrics provide quantitative measures of accuracy and fluency, enabling a more comprehensive analysis of the translation quality.

\subsubsection{Accuracy}
To evaluate the accuracy of translated sentences, we utilize the LASER model \cite{heffernan-etal-2022-bitext}, which excels in capturing semantic meaning and supports a wide range of languages except for Ibibio, which is not in the list of 147 languages that the LASER encoder supports. We employ this model to compute the embeddings of the translated sentences and their corresponding English sentences. To compare the embeddings, we compute the cosine distance, which serves as a measure of semantic similarity between the translations and the source English sentences. A higher cosine similarity indicates a stronger alignment in semantic meaning, suggesting higher accuracy in the translation process.

In addition to evaluating the accuracy between the translated sentences and their corresponding English sentences, we also examine the similarity between the translations generated through our storyboard-based method and the text translation method. To measure whether the two methods result in translations with comparable accuracy, we calculate the cosine similarity between embeddings of the translated sentences. The cosine similarity provides an indication of how similar the translations are in terms of their semantic meaning. While we do not expect a perfect similarity between the two methods' translations, we aim for a comparable level of meaning that allows for fluency trade-off.

\subsubsection{Fluency}
Fluency in translation involves the successful use of vocabulary and sentence structure to convey meaning effectively in the target language. When translationese phenomena occur, patterns emerge that can impact the fluency of translated sentences. Previous studies \cite{vanmassenhove-etal-2021-machine, bizzoni-etal-2020-human} have highlighted two key factors in fluency: lexical diversity and syntactic complexity.

\textbf{Lexical Diversity} To assess the lexical diversity or vocabulary richness of the translated sentences, we employ the Measure of Textual Lexical Diversity (MTLD) metric \cite{mccarthy2005assessment}. Conceptually, MTLD reflects the average number of words in a row for which a certain TTR (Type-Token Ratio) is maintained, specifically in this analysis, following \citet{mccarthy_mtld_2010} we use TTR of $0.72$. To generate a score, MTLD calculates the TTR for increasingly longer parts of the sample. Every time the TTR drops below the predetermined value, a count (called the factor count) increases by 1, and the TTR evaluations are reset. The algorithm resumes from where it had stopped, and the same process is repeated until the last token of the language sample has been added and the TTR has been estimated. Then, the total number of words in the text is divided by the total factor count. Subsequently, the whole text in the language sample is reversed and another score of MTLD is estimated. The forward and the reversed MTLD scores are averaged to provide the final MTLD estimate.

Intuitively the MTLD value can be seen as the average number of words required for the text to reach a point of stabilization.

\textbf{Syntactic Complexity:}
To evaluate syntactic complexity, we leverage an off-the-shelf language model trained on the languages used in the study, specifically AfroXLMR-base \cite{alabi-etal-2022-adapting}---an adaptation of XLM-R to 17 African languages, including Hausa, Swahili and \yoruba. To estimate the POS perplexity \cite{bizzoni-etal-2020-human}, we trained a Part-Of-Speech (POS) model by fine-tuning the AfroXLMR language model on MasakhaPOS dataset~\citep{DioneMasakhanPOS}--- a large-scale POS dataset for 20 African languages, including all focus languages except for Ibibio. POS perplexity measures the difficulty in predicting the sequence of POS tags in a sentence. Higher perplexity values indicate greater syntactic complexity. Perplexity is defined as the exponentiation of the entropy:
\begin{equation}
    2^{H(p)}=2^{-\sum_x p(x) \log_2 p(x)}
\end{equation}
Where $p(x)$ is the probability of the predicted POS.

\subsection{Results}
\label{sec:results}

\subsubsection{Accuracy Evaluation}

\begin{table}[htbp]
\centering
\begin{tabular}{lrrr}
\hline
\textbf{Language} & \textbf{Storyboard} & \textbf{Text} & \textbf{Both} \\ \hline
Hausa & 21.67\% & 78.33\% & 0\% \\ 
Swahili & 14\% & 64\% & 22\% \\ 
\yoruba & 7.67\% & 68.67\% & 23.67\% \\ 
Ibibio & 18.33\% & 79.67\% & 2\% \\ \hline
\end{tabular}
\caption{Human Evaluation Results for Accuracy. More annotators choose text translations as more accurate across languages}
\label{tab:human-eval-ade}
\end{table}

\begin{table}
    \centering
    \begin{tabular}{lcc}
    \hline
        \textbf{Language} & \textbf{Storyboard} & \textbf{Text}\\ \hline
         Hausa & $0.64 \pm 0.14$ & $0.74 \pm 0.13$ \\
         Swahili & $0.58 \pm 0.11$ & $0.69 \pm 0.10$\\
         \yoruba & $0.64 \pm 0.13$ & $0.74 \pm 0.12$ \\
         \hline
    \end{tabular}
    \caption{The Average Cosine Similarity Scores between English and Translated Sentences' LASER Embeddings. Across languages, text translations have higher semantic similarity to source English sentences, although the scores still fall within each method's standard deviation}
    \label{tab:laser-results}
\end{table}

The accuracy evaluation results in Table \ref{tab:human-eval-ade} show a clear preference among human annotators for text translations in terms of accuracy across all languages. This is consistent with the expectation that text translations, typically done by professional translators, would be more semantically aligned with the source sentences.

The cosine similarity values between LASER embeddings of the translated sentences and their corresponding English sentences, as shown in Table \ref{tab:laser-results}, further support this observation. Text translations generally exhibit higher semantic similarity to the source English sentences. However, it's important to note that while the cosine similarity values for storyboard translations are lower, they are still within the standard deviation range of the text translations, suggesting that the semantic content is not drastically different.

The cosine similarity between translations from the two methods, presented in Table \ref{tab:laser-comparable}, indicates that the semantic content of translations from both methods is relatively comparable, even if they are not identical.

\begin{table}
    \centering
    \begin{tabular}{lc}
        \hline
        \textbf{Language} &  \textbf{Cosine Similarity}\\ \hline
        Hausa & $0.63 \pm 0.19$ \\
        Swahili & $0.62 \pm 0.17$ \\
        \yoruba & $0.59 \pm 0.18$ \\
        \hline
    \end{tabular}
    \caption{The Average Cosine Similarity Scores between text translations and storyboard-based translations' LASER embeddings.}
    \label{tab:laser-comparable}
\end{table}

\subsubsection{Fluency Evaluation}
\begin{table}[htbp]
\centering
\scriptsize
\begin{tabular}{lrrrr}
\hline
\textbf{Language} & \textbf{Storyboard} & \textbf{Text} & \textbf{Both} & \textbf{p-value}\\ \hline
Hausa & 60\% & 39.67\% & 0.33\% & 0.0002 \\ 
Swahili & 47.67\% & 41.33\% & 11\% & 0.11 \\ 
\yoruba & 34\% & 18.6\% & 47.33\% & 0.0008\\ 
Ibibio & 36\% & 26\% & 38\% & 0.01\\ \hline
\end{tabular}
\caption{Human Evaluation Results for Fluency. More annotators choose storyboard translations as more fluent across languages. P-Value is for the null hypothesis of the annotators choosing by random.}
\label{tab:human-eval-flu} 
\end{table}

\begin{table}
    \centering
    \begin{tabular}{lcc}
    \hline
        \textbf{Language} & \textbf{Storyboard} & \textbf{Text}\\ \hline
        Ibibio & $ 8.08 \pm 17.06$ & $ 7.27 \pm 18.35 $ \\
        Hausa & $ 7.8 \pm 16.77$ & $ 11.41 \pm 23.48 $ \\
        \yoruba & $ 16.12 \pm 30.05$ & $ 14.16 \pm 29.84 $  \\
        Swahili & $ 6.83 \pm 19.64$ & $ 5.03 \pm 16.61 $ \\
         \hline
    \end{tabular}
    \caption{MTLD of Translated Sentences. Across languages, except for Hausa, storyboard translations have higher average MTLD scores}
    \label{tab:mtld-results}
\end{table}

The fluency evaluation in Table \ref{tab:human-eval-flu} suggests a preference among human annotators for storyboard translations in terms of fluency across all languages. This indicates that while storyboard translations might not always capture the exact semantic content of the source, they tend to produce more naturally flowing sentences in the target languages.

The MTLD scores in Table \ref{tab:mtld-results}, despite having a high standard deviation, provide further evidence for this observation. Except for Hausa, storyboard translations generally exhibit greater lexical diversity, which is often associated with more natural and fluent sentences.

The POS perplexity values in Table \ref{tab:pos-results} offer insights into the syntactic complexity of the translations. Lower POS perplexity values typically indicate more natural sentence structures in the target language. The results suggest that, except for Swahili, storyboard translations tend to produce more naturally structured sentences than text translations.

In summary, while text translations are generally more accurate and semantically aligned with the source, storyboard translations offer advantages in terms of fluency and naturalness of the produced sentences.

\begin{table}
    \centering
    \begin{tabular}{lcc}
    \hline
        \textbf{Language} & \textbf{Storyboard} & \textbf{Text}\\ \hline
        
         Hausa & $ 6.68 $&$ 49.42 $ \\
         Swahili & $ 55.32 $&$ 14.6 $\\
         \yoruba & $ 6.49 \times 10^2 $ &$ 7.93 \times 10^7 $ \\
         \hline
    \end{tabular}
    \caption{POS Perplexity of Translated Sentences. Across languages, except for Swahili, the storyboard translations have lower POS perplexity, indicating more natural translations in terms of sentence structure than the text translations}
    \label{tab:pos-results}
\end{table}

\section{Discussion}

\subsection{Comparison of Accuracy Evaluation}

The human accuracy evaluation results (Table \ref{tab:human-eval-ade}) reveal a clear preference for text translation over our proposed method. Human evaluators consistently rated text translation as more accurate compared to our method. This preference aligns with the linguistic intuition that text translations, performed by professional translators, tend to produce more accurate translations. Similarly, looking at the cosine similarities between the source English sentences and the translations (Table \ref{tab:laser-results}), we can see that text translations have higher cosine similarities to the English sentences, indicating higher semantic similarities. This observation is consistent with human evaluators' preference for accuracy. 

However, when comparing with the cosine similarities between our method's (storyboard) translations and the text translations (Table \ref{tab:laser-comparable}), we found that the cosine similarity scores were not significantly lower. This suggests that the information conveyed through our storyboard translations remains comparable to that of text translations, albeit with a slightly lower degree of semantic similarity. Despite the storyboard translations not matching the accuracy of the text translations perfectly, our storyboard-based data collection method still provides translations that are reasonably accurate and comparable.

\subsection{Comparison of Fluency Evaluation}

The human fluency evaluation presents a contrast to the accuracy evaluations. Human evaluation 
results reveal a clear preference for our storyboard translations over text translations in terms of fluency (Table \ref{tab:human-eval-flu}).

In analyzing the fluency metrics, looking at the Measure of Textual Lexical Diversity (MTLD), %
our storyboard translations consistently yielded higher MTLD scores compared to text translations for Ibibio, \yoruba, and Swahili (Table \ref{tab:mtld-results}). This suggests that our storyboard-based data collection method can capture a wider range of vocabulary and exhibit greater lexical diversity, enhancing the fluency of the resulting translations in these languages. 

We also evaluated the syntactic complexity of the translations using the Part-Of-Speech (POS) perplexity metric, which provide insights into the syntactic intricacy and sentence structure of the translations. Looking at POS perplexity, our storyboard translations exhibited lower POS perplexity values for Hausa and \yoruba compared to text translations (Table \ref{tab:pos-results}). This suggests that our storyboard-based data collection method 
achieves a more natural sentence structure in these languages. 

\subsection{Reflections on the Storyboard Approach}
The storyboard approach, while pioneering, calls for an in-depth examination of its intrinsic strengths and limitations, especially when contrasted against conventional text translation techniques.

\textbf{Enhanced Fluency} As underscored by our findings, the storyboard technique frequently yields translations perceived as more fluent by human evaluators. This can be ascribed to the more spontaneous elicitation process, where native speakers articulate visual stimuli, culminating in more innate sentence constructs.

\textbf{Lexical Diversity}: The method appears to encapsulate a broader lexicon, potentially enriching the translations with diverse linguistic expressions.

\textbf{Accuracy Concerns} The storyboard approach, while fluent, occasionally compromises on accuracy. This is seen in both human evaluations and cosine similarity metrics. The omission or modification of certain nuances, particularly named entities, can influence the perceived precision of the translations. We have a few suggestions on how to improve this. Firstly, refining the storyboards to provide clearer context, especially regarding named entities, can help translators capture essential details. Secondly, implementing post-processing techniques can further align the translations with the source content. By incorporating correct named entities from the source and removing extraneous information, we can achieve a better balance between fluency and accuracy. Finally, introducing temporal separation between viewing the source and translating can help translators produce more natural translations, less influenced by the source language structure. By integrating these strategies, we aim to enhance the accuracy of the storyboard-based approach without compromising its inherent fluency.

\textbf{Feasibility for Complex Texts} The viability of the method for translating intricate and detailed texts remains a point of contention. Storyboards, by their visual essence, might fall short in capturing the depth and nuances of elaborate narratives or technical documents.

\section{Future Work}

\begin{figure}[t]
\includegraphics[scale=0.4]{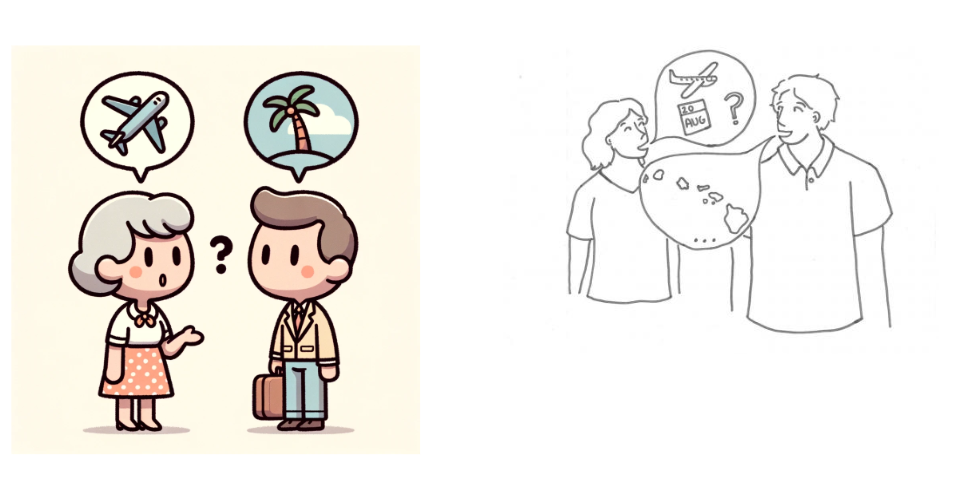}
\centering
		\caption{Comparison between the DALLE-3 generated storyboard (left) and the manually designed storyboard (right). The similarities highlight the potential of generative AI in automating the storyboard creation process}
		\label{fig:storyboard_comparison}
		\vspace{-1.5em}
\end{figure}

The storyboard approach, while innovative, faces challenges in terms of the resource-intensive nature of creating detailed storyboards. A promising solution lies in the integration of generative AI models, such as DALLE-3. Our initial exploration with DALLE-3 to recreate a manually designed storyboard showed potential in automating the storyboard creation process, as evidenced by the comparison in Figure \ref{fig:storyboard_comparison}. However, inconsistencies in DALLE-3's generated characters and the challenge of conveying complex messages visually remain areas for improvement. As generative AI models evolve, there's potential for more sophisticated storyboard generation. Collaborative efforts between linguists, artists, and AI researchers could lead to hybrid methodologies that merge manual design with AI-driven automation. Other future work could consider expanding the complexity of messages captured in storyboard, and improving the accuracy of the storyboard.
% \subsection{Expanding the Storyboard Spectrum}

% Addressing the feasibility concerns of the storyboard approach requires the creation of more detailed and nuanced storyboards. This expansion can capture intricate narratives more effectively, especially with the potential assistance of advanced AI models.

% \subsection{Bias Mitigation and Addressing Complex Messages}

% Ensuring the unbiased and comprehensive representation of narratives in storyboards is crucial. Collaborative storyboard creation, involving a diverse group of contributors, can help in identifying and mitigating biases. Additionally, the challenge of encapsulating complex or abstract concepts visually remains an area for exploration and innovation.

% In summation, while the storyboard approach proffers a fresh vantage point on translation data collection, it's pivotal to approach it with a balanced perspective, recognizing its potential and its pitfalls.  As with any nascent method, continuous refinement and feedback will be instrumental in unlocking its full potential.

\section{Conclusion}
% We proposed a novel approach to low-resource translation data collection using storyboards and image-based translation. Through human evaluations and automated accuracy and fluency metrics, we observed that while text translations showed a higher degree of translation accuracy, our proposed method demonstrated comparable accuracy and superior fluency in several languages. We observed that the incorporation of image-based translation enhanced lexical diversity and more natural sentence structures, resulting in more fluent translations. More generally, these findings have implications for the development of data collection systems that can mitigate the translationese phenomena. The potential applications and impact of our approach include improvement in performance of machine translation task and other tasks that will benefit from better cross-lingual training data and transfer. %aiding language learners, and enhancing cross-cultural communication. 
We introduced an alternative method for gathering translation data in low-resource settings using storyboards and visual-based translation techniques. Our findings, derived from both human assessments and automated metrics for accuracy and fluency, revealed that while traditional text translations excel in accuracy, our method offers a balance of accuracy and heightened fluency across multiple languages. Notably, the use of visual-based translation enriched lexical variety and fostered more organic sentence formations, leading to translations that resonate more naturally. these findings have implications for the development of data collection systems that can mitigate the translationese phenomena. The potential applications and impact of our approach include improvement in performance of machine translation task and other tasks that will benefit from better cross-lingual training data and transfer. %aiding language learners, and enhancing cross-cultural communication. 

\section*{Limitations}
While our study focuses on and observes the effectiveness of the storyboard-based data collection method and its potential to mitigate the translationese effect, it is important to acknowledge certain limitations. 

Firstly, our study focused on four low-resource African languages, namely Swahili, \yoruba, Hausa, and Ibibio. While we also contribute to the development of parallel resources for these languages (including the first-ever parallel resource for Ibibio in non-religious domain), our analysis and findings on the translationese phenomena in the resulting dataset may not necessarily generalize to other languages or language families. Therefore, caution should be exercised when applying the results to different linguistic contexts. In addition, our study specifically focused on the use of storyboards to collect data in these low-resource languages. The effectiveness of this method may vary in different data collection scenarios, such as with different types of visual stimuli or in languages with different linguistic characteristics.

Secondly, the number of annotators and the size of the dataset used in our study were limited. Although efforts were made to mitigate bias and ensure reliability through multiple annotators, a larger sample size could provide more robust and representative results. In addition, the human evaluation process involves subjective judgments made by human annotators. Individual preferences and biases may influence the evaluation results. While we attempted to minimize bias by using multiple annotators and consistent sentence pairings, the subjective nature of the evaluation should be considered. 

Thirdly, in translating, the fluency and naturalness of the translated sentences can be influenced by various external factors, such as the annotators' language proficiency, cultural background, and familiarity with the subject matter. While efforts were made to select skilled annotators, these factors may still have an impact on the results.

Furthermore, the storyboard-based data collection method, while innovative, is inherently more challenging compared to traditional text translation. This method involves a more complex setup for data collection, as participants need to be effectively oriented to provide translations based on visual stimuli rather than textual source material. These logistical and financial considerations make the method potentially less scalable and more expensive than traditional approaches, especially for larger datasets or multiple languages.

Lastly, our evaluation mainly focused on accuracy and fluency, and utilized metrics such as cosine similarity, lexical diversity, and syntactic complexity. While these metrics provide some measures of translationese, they may not capture all aspects of translation quality. Additional metrics or qualitative assessments could further enhance the evaluation.

\section*{Ethics Statement}
%Scientific work published at ACL 2023 must comply with the ACL Ethics Policy.\footnote{\url{https://www.aclweb.org/portal/content/acl-code-ethics}} We encourage all authors to include an explicit ethics statement on the broader impact of the work, or other ethical considerations after the conclusion but before the references. The ethics statement will not count toward the page limit (8 pages for long, 4 pages for short papers).
The data collection process adhered to standard guidelines. Informed consent was obtained from all participants involved in the study, and they were fully informed about the purpose of the research, their rights, and how their data would be used. Participants were assured of the anonymity and confidentiality of their information throughout the data collection process.

\section*{Acknowledgements}

This work is supported in part by the DARPA HR001118S0044 (the LwLL program) and the Indonesia-US Research Collaboration in Open Digital Technology grant funded by the Indonesian Ministry of Education,
Culture, Research, and Technology. The U.S. Government is authorized to reproduce and distribute reprints for Governmental purposes. The views and conclusions contained in this publication are those of the authors and should not be interpreted as representing official policies or endorsements of DARPA or the U.S. Government.

\section{Bibliographical References}\label{sec:reference}
\bibliographystyle{lrec-coling2024-natbib}
\bibliography{lrec-latex/lrec-coling2024-example}%, 
%lrec-latex/anthology}

% \section{Language Resource References}
% \label{lr:ref}
%\bibliographystylelanguageresource{lrec-coling2024-natbib}
%\bibliographylanguageresource{languageresource}

\end{document}